\documentclass{article}

\usepackage{microtype}
\usepackage{graphicx}
\usepackage{subfigure}
\usepackage{booktabs} 
\usepackage{enumitem}
\usepackage{xspace}

\usepackage{listings}

\usepackage{hyperref}

\usepackage[accepted]{icml2023}

\usepackage{amsmath}
\usepackage{amssymb}
\usepackage{mathtools}
\usepackage{amsthm}
\usepackage{array}
\usepackage{bm}

\usepackage[T1]{fontenc}

\usepackage[utf8]{inputenc}

\usepackage[capitalize,noabbrev]{cleveref}
\usepackage{color, colortbl, soul}

\usepackage{xcolor}

\definecolor{darkgreen}{HTML}{005e19}
\definecolor{darkblue}{HTML}{240394}
\definecolor{lightblue}{HTML}{edf6ff}
\definecolor{lightpink}{HTML}{ffeded}
\definecolor{lightgreen}{HTML}{f2ffed}
\definecolor{fig1orange}{HTML}{e8612a}
\definecolor{fig1blue}{HTML}{38a3e4}
\sethlcolor{lightpink}

\definecolor{green1}{HTML}{0b5400}
\definecolor{orange1}{HTML}{f3905c}
\definecolor{blue1}{HTML}{027db5}
\definecolor{pink1}{HTML}{ff7a7a}

\usepackage{fdsymbol}
\usepackage{multirow}
\newcommand{\sclub}{$\textcolor{blue1}{\clubsuit}$}
\newcommand{\sspade}{$\textcolor{green1}{\spadesuit}$}
\newcommand{\sdiamond}{$\textcolor{orange1}{\vardiamondsuit}$}
\newcommand{\sheart}{$\textcolor{pink1}{\varheartsuit}$}

\newcommand{\code}[1]{\texttt{#1}}

\newcommand{\expblue}[1]{\textcolor{darkblue}{\textbf{\small{\code{#1}}}}}

\theoremstyle{plain}

\theoremstyle{definition}

\theoremstyle{remark}

\usepackage[textsize=tiny]{todonotes}

\icmltitlerunning{A song of ice and fire: analyzing textual autotelic agents in ScienceWorld}

\begin{document}

\twocolumn[
\icmltitle{A Song of Ice and Fire: Analyzing Textual Autotelic Agents in ScienceWorld}



\icmlsetsymbol{equal}{*}

\begin{icmlauthorlist}
\icmlauthor{Laetitia Teodorescu}{inria}
\icmlauthor{Xingdi Yuan}{msr-mtl}
\icmlauthor{Marc-Alexandre C\^ot\'e}{msr-mtl}
\icmlauthor{Pierre-Yves Oudeyer}{inria}
\end{icmlauthorlist}

\icmlaffiliation{inria}{Flowers Team, Inria, France}
\icmlaffiliation{msr-mtl}{Microsoft Research, Montreal, Canada}

\icmlcorrespondingauthor{Laetitia Teodorescu}{laetitia.teodorescu@inria.fr}

\icmlkeywords{Machine Learning, Natural Language Processing, Text Environments, Automatic Exploration, Autotelic Agents}

\vskip 0.3in
]

\newcommand{\chain}{goal-chain\xspace}
\newcommand{\explore}{go-explore\xspace}
\newcommand{\Chain}{Goal-chain\xspace}
\newcommand{\Explore}{Go-explore\xspace}
\newcommand{\SP}{$\mathcal{SP}$\xspace}
\newcolumntype{P}[1]{>{\centering\arraybackslash}p{#1}}



\printAffiliationsAndNotice{\icmlEqualContribution} 


\begin{abstract}
Building open-ended agents that can autonomously discover a diversity of behaviours is one of the long-standing goals of artificial intelligence. This challenge can be studied in the framework of autotelic RL agents, i.e. agents that learn by selecting and pursuing their own goals, self-organizing a learning curriculum. Recent work identified language as a key dimension of autotelic learning, in particular because it enables abstract goal sampling and guidance from social peers for hindsight relabelling. Within this perspective, we study the following open scientific questions: \textit{What is the impact of hindsight feedback from a social peer (e.g. selective vs. exhaustive)? How can the agent learn from very rare language goal examples in its experience replay? How can multiple forms of exploration be combined, and take advantage of easier goals as stepping stones to reach harder ones?} To address these questions, we use ScienceWorld, a textual environment with rich abstract and combinatorial physics. We show the importance of selectivity from the social peer's feedback; that experience replay needs to over-sample examples of rare goals; and that following self-generated goal sequences where the agent's competence is intermediate leads to significant improvements in final performance.\footnote{The anonymized code can be found at \href{https://anonymous.4open.science/r/textworld-explore-C015/README.md}{this url}.}
\end{abstract}


\section{Introduction}
\label{sec:intro}

We are interested in the problem of building and training open-ended autonomous agents, exploring on their own and mastering a wide diversity of tasks once trained. This can be approached within the autotelic reinforcement learning framework \cite{colas2022autotelic}. An autotelic agent (auto-telos, one’s own goals) is an intrinsically-motivated, goal-conditioned agent equipped with a goal-sampler that uses the agent’s previous experience to propose goals for learning and exploration. This goal sampler allows the agent to use previously mastered skills as stepping stones to achieve new ones, and to form a self-curriculum for exploration. This developmental framework is general and is linked to goal-exploration processes like Go-explore \cite{ecoffet2021first} and adversarial goal generation \cite{florensa2018automatic, campero2020learning}. Autotelic agents have already been shown to efficiently explore sensorimotor spaces \cite{pere2018unsupervised} and to be able to build their own goal curriculum using learning-progress-based task sampling \cite{pmlr-v97-colas19a}.

Recent work has shown the potential of language to drive autotelic learning \cite{colas2022language}, both due to its compositional structure and its ability to convey cultural knowledge. For example, post-episode language feedback by a social peer (\SP) can be internalized to identify and imagine future relevant goals \cite{colas2020language}, acting as a cognitive tool \cite{Clark2006-CLALEA-3}. It has also been shown to help scaffold the agent's exploration through abstraction \cite{mu2022improving}. It can be reused once the agent is trained for instruction-following \cite{colas2022language}. Exploring in language space directly is akin to learning to plan at a higher, abstract level; the skills learned at this level can be executed by lower-level modules in an embodied environment \cite{shridhar2020alfworld, saycan2022arxiv, huang2022language}. Additionally, language conveys our morals and values, and be used as a tool to align autotelic agents towards human preferences \cite{sigaud2021towards}.

\begin{figure*}
\centering
\includegraphics[width=450pt]{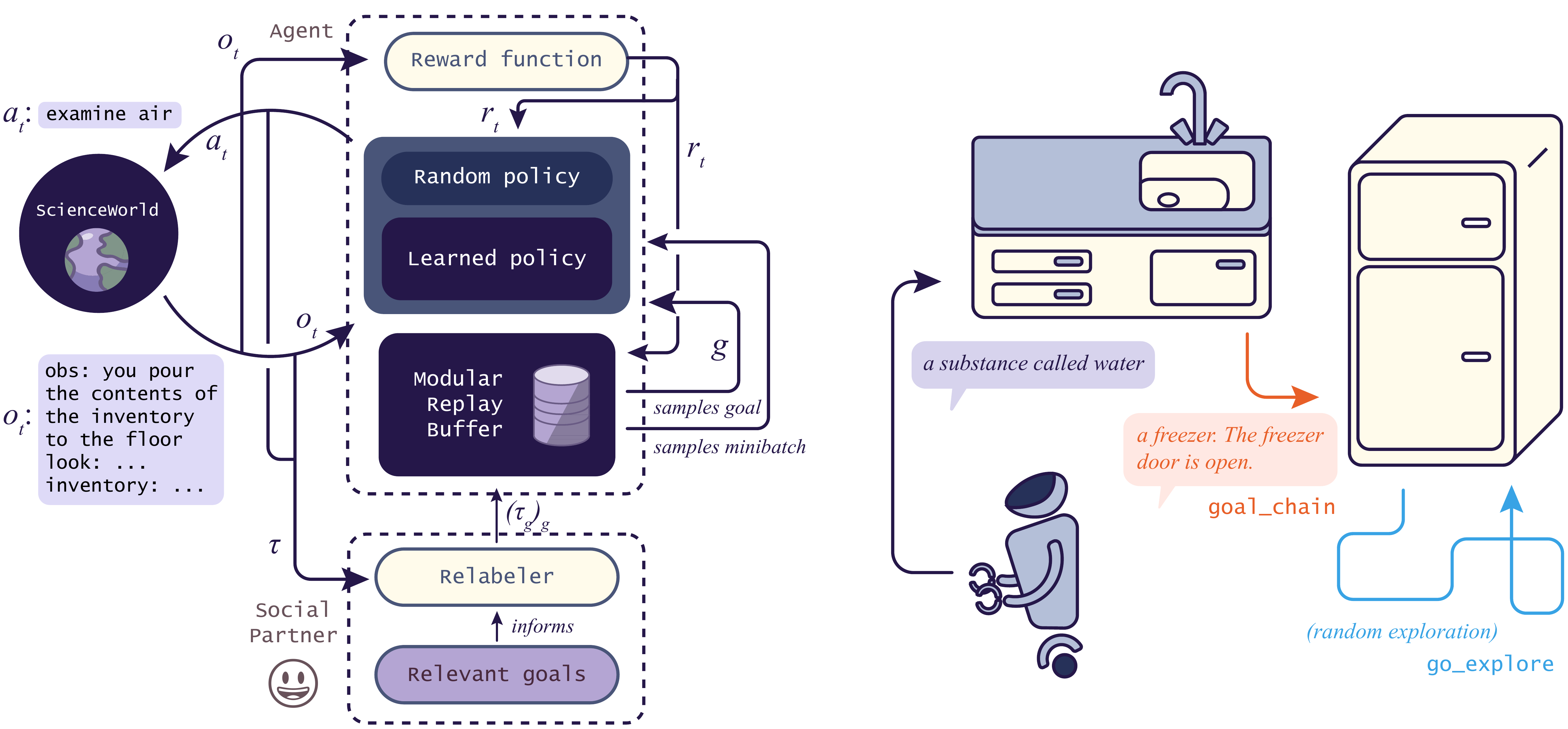}
\caption{Left: overview of the autotelic agent architecture. At the beginning of an episode, a goal $g$ is sampled from the list $\mathcal{G}_a$ maintained in the modular replay buffer. At each timestep $t$, ScienceWorld emits an observation $o_t$ (see Section \ref{sec:sw}). The agent combines $o_t$ and $g$ to decide on an action $a_t$. $o_t$ and $g$ are also used by the agent to compute the reward $r_t$. When the episode ends ($r_t \neq 0$ or $t = T$), either the environment is reset, a new goal is sampled or random exploration steps are taken.  Right: overview of different exploration methods. The agent is conditioned on a goal and tries to achieve it. \textcolor{fig1orange}{In the \chain configuration}, on achieving a goal or at the end of the allowed $T$ timesteps the agent samples a new goal with a 0.5 probability. \textcolor{fig1blue}{In the \explore configuration}, on achieving a goal or at the end of the $T$ timesteps the agent performs 5 timesteps of random exploration.}
\label{fig:overview}
\end{figure*}

The common topic in previous work \cite{colas2020language, mirchandani2021ella, mu2022improving} on language-based autotelic agents has been to study various forms of exploration mechanisms, some of them relying on language. However, these works did not investigate how to deal with the exploration challenges posed by linguistic spaces themselves, featuring immense action and goal spaces. More precisely, \textit{how specific should \SP's feedback be? How to deal with very hard goals that will be very rarely seen compared to easy ones? How to use easy goals as stepping stones to achieve hard ones?} These are the questions we focus on in this work. For studying this, we place ourselves in ScienceWorld \cite{wang2022scienceworld}, a text world \cite{cote2019textworld} with very rich dynamics (thermodynamics, biology, electricity) allowing for complex goals such as freezing or boiling water, which requires getting water first, thus defining an optional dependency amongst goals (ScienceWorld is rich enough that goals can be accomplished in multiple ways).

To tackle these challenges, we identify the main drivers of discovery in autotelic agents. In general, there are four ways in which they can discover novel things through goal exploration in an environment. The agent can \textbf{discover from failure}: it can target a goal it misses, and the social peer will relabel this trajectory with the actually achieved goals (if any), an idea exploited in hindsight experience replay (HER) \cite{andrychowicz2017hindsight}. The agent can \textbf{discover from babbling}: it is equipped with standard RL exploration mechanisms, such as stochastic policies or epsilon-greedy action sampling. This will induce randomness in goal-conditioned trajectories and allow the agent to stumble onto novel states. The agent can \textbf{discover from stepping-stone exploration}: after a goal has been reached or the episode has timed out, the agent can perform random actions or follow another goal from there, a process first investigated by Go-explore methods. Here exploration bonuses (like pseudo-count rewards \cite{bellemare2016unifying}) can be used to make this process more efficient. And finally, the agent can \textbf{discover from imagination}: combine known goals to create novel ones, leading to the discovery of novel states and affordances (which can in turn be reused as goals in further exploration). To tackle the exploration challenges posed by large linguistic spaces and nested goals, we especially focus on the first three drivers of discovery. In particular, we show how different methods for learning from further exploration interact together to help the agent navigate the goal hierarchy.

In this work, we study specific challenges posed by autotelic exploration in large language spaces:

\begin{enumerate}[wide, labelwidth=!, labelindent=0pt]
\item \textit{How should the \SP provide hindsight feedback (relabeling) to the agent in very large linguistic spaces? Should it be selective or exhaustive?}
We show the social peer must give targeted hindsight feedback to the agent to avoid populating the replay buffer with a too wide diversity of detailed goals that prevents making non-trivial discoveries.
\item \textit{In the presence of goals with very different difficulty and occurrences, what is the influence of different goal sampling distributions on the efficiency of learning diverse and complex goals?} We show that the agent needs to bias replay transition sampling towards transitions in trajectories where rare, hard goals are accomplished.
\item \textit{How do methods for learning from stepping-stone exploration influence learning in the goal hierarchy?} We find that sampling goal sequences according to the agent's estimated intermediate competence, and then exploring randomly, significantly improves aggregate competence and reduces variance over seeds.
\end{enumerate}

\section{Problem setting}
\label{sec:scienceworld}


%
%

\begin{figure}[t!]
\begin{center}
\scriptsize
\setlength{\tabcolsep}{3pt}
\begin{tabular}{p{0.98\linewidth}} 
\toprule
    \textbf{ScienceWorld Game} \\
\midrule
\textit{> look} \\
\sethlcolor{lightpink}
\hl{\textbf{Kitchen}} \\
\hl{This room is called the kitchen. In it, you see:

\:\:\:\:the agent\\
\:\:\:\:a substance called air\\
\:\:\:\:a counter. On the counter is: \\
\:\:\:\:\:\:\:\:a bowl (containing a red apple, a banana, an orange, a potato),\\
\:\:\:\:a freezer. The freezer door is closed.\\
\:\:\:\:a fridge. The fridge door is closed.\\
\:\:\:\:a glass jar (containing a substance called sodium chloride)\\
\:\:\:\:a sink, which is turned off. In the sink is: nothing.\\
\:\:\:\:a substance called soap\\
\:\:\:\:a \textbf{stove}, which is turned off. On the \textbf{stove} is: nothing.\\
\:\:\:\:a table. On the table is: a \textbf{glass cup} (containing nothing). }\\
\\

\textit{> inv} \\
\sethlcolor{lightblue}
\hl{In your inventory, you see:\\
\:\:\:\:an orange.}\\
\\
\textit{> pick up \textbf{glass cup}} \\
\sethlcolor{lightgreen}
\hl{You move the \textbf{glass cup} to the inventory.} \\
\\

\textit{> move \textbf{glass cup} to sink} \\
You move the \textbf{glass cup} to the sink. \\
\\

\textit{> activate sink} \\
The sink is now activated. \\
\\

\textit{> look in sink} \\
Inside the sink is: \\

\:\:\:\:a \textbf{glass cup} (containing a \textbf{substance called water}) \\

You also see: A drain, which is open. \\
\\
\textit{> move \textbf{glass cup} to \textbf{stove}} \\
You move the \textbf{glass cup} to the \textbf{stove}. \\

\bottomrule

\end{tabular}
\caption{\footnotesize A ScienceWorld trajectory illustrating an agent puts a glass cup full of water on the stove. 
For accessibility purpose, we omit repetitive text (agent receives \code{obs}, \code{look}, and \code{inv} at every step, as described in Section~\ref{sec:sw}).
Italic and boldface denote \textit{action} and \textbf{relevant object}. 
}
\label{fig:scienceworld-trajectory-example}
\end{center}
\vspace{-6mm}
\end{figure}

We are interested in studying the behavior of autonomous agents that freely explore their environments to uncover their possibilities. Such agents are especially needed in environments that are reward-less, or that have sparse or deceptive rewards.

\subsection{Definitions}
\paragraph{Reward-less POMDP} Formally, we define a reward-less partially observable Markov decision process (POMDP)~\cite{SuttonRL} as $(S, \mathcal{A}, \mathcal{T}, \Omega, \mathcal{O})$, where $S$ is the state space, $\mathcal{A}$ is the action space, $\mathcal{T}$ is the transition function, $\Omega$ is the observation space, and $\mathcal{O}$ is the observation function. We also define a trajectory as a sequence of state-action pairs and is represented by $\tau = [(s_0, a_0), \dots, (s_t, a_t), \dots, (s_T, a_T)]$ where $t$ is a timestep and $T$ the length of the trajectory.

\paragraph{Autotelic agents} In this work, we consider a certain kind of autonomous agents that are driven by an intrinsically-motivated goal-exploration process: autotelic agents. These agents operate without external reward by iteratively targeting goals and trying to reach them, using their own internal goal-achievement (reward) function to measure success. In the process, they observe their own behavior and learn from it. We define $\mathcal{G}_a \subseteq \mathcal{G}$ as the subset of goals experienced by the agent so far during its learning process. We additionally define the agent's goal-conditioned internal reward function as $R: S \times A \times \mathcal{G} \rightarrow \{0, 1\}$.

Ideally, we would want to maximize the agent's performance over the entire goal space $\mathcal{G}$, i.e., to find the goal-conditioned policy $\pi$ that maximizes the expected sum of internal rewards on all goals.
However, these goals are not known in advance and have to be discovered by the agent through structured exploration. This means that this objective cannot be computed and used directly by the agent. Rather, it can be used \textit{a posteriori} by the experimenter as a measure to characterize what the agent discovered and learned. 

\paragraph{Social peer} Since $\mathcal{G}$ can be quite large, it would be desirable to guide the exploration without forcing it. We consider the agent interacts with a social peer (\SP) that gives feedback on the agent's trajectories, i.e., the \SP gives a list of achieved goals at the end of an episode (this list may not be exhaustive and reflects a model of relevance from the perspective of the \SP). Formally, we define the social peer as $\mathcal{SP}: (S \times A)^T \rightarrow (\mathcal{G}_{\mathcal{SP}})^{m}$ which takes in a trajectory and outputs a set of $m$ goals accomplished within this trajectory, where $\mathcal{G}_{\mathcal{SP}} \subseteq \mathcal{G}$ are goals relevant to \SP. 
Then, those accomplished goals can be added to the agent's discovered goals $\mathcal{G}_a$.
In principle, goal relabelling can be implemented in many ways, including leveraging pre-trained large language models.
In this study, for simplicity, the \SP labels objects presented in trajectories as goals (see Section~\ref{sec:sw}).

In practice, we are maximizing the agent's performance to achieve the goals it has discovered:
\begin{equation}
\sum_{g \in \mathcal{G}_a} \mathop{\mathbb{E}}_{\tau \sim \pi(.|g)} \Big[ \sum_{(s_t, a_t) \in \tau} \gamma^t \: R(s_t, a_t, g) \Big]
\end{equation}
while $\mathcal{G}_a$ converges towards $\mathcal{G}_{\mathcal{SP}}$ as trajectories gets relabelled by \SP. In which, $\gamma$ denotes the discount factor.

\subsection{ScienceWorld: a text-based environment}
\label{sec:sw}

ScienceWorld\footnote{An online demo for ScienceWorld is available at \href{https://github.com/allenai/ScienceWorld\#demo-and-examples}{github.com/allenai/ScienceWorld\#demo-and-examples}}~\cite{wang2022scienceworld} is one of the text world frameworks~\cite{jansen2022survey}, coming with procedurally-generated household environments and an associated elementary-school science related benchmark. 
Unlike many other interactive environments that facilitates RL research, in ScienceWorld, the observation space $\Omega$ and the action space $\mathcal{A}$ are textual. 
Specifically, at every timestep, the \textbf{observation} consists of three channels (please refer to Figure \ref{fig:scienceworld-trajectory-example} for an in-game example):\\
\sethlcolor{lightgreen}
$\bullet$ \code{obs}: a raw observation describing the immediate effect of the agent's actions. We show an example in Figure~\ref{fig:scienceworld-trajectory-example}, highlighted in \hl{green}. \\
\sethlcolor{lightpink}
$\bullet$ \code{look}: the description of the agent's surrounding. It is composed of a textual rendering of the underlying object tree, with receptacle-content hierarchy. We show an example in Figure~\ref{fig:scienceworld-trajectory-example}, highlighted in \hl{pink}. \\
\sethlcolor{lightblue}
$\bullet$ \code{inv}: a description of the agent's inventory. We show an example in Figure~\ref{fig:scienceworld-trajectory-example}, highlighted in \hl{blue}. 

The (text-based) \textbf{action} space is combinatorial and extremely large. Actions are templated phrases built by composing predicates (from a list of 25 possible unary or binary predicates) with compatible attributes, leading to 200k possible actions at each timestep on average.
To alleviate this issue, ScienceWorld provides a shorter list of valid actions $\mathcal{A}_t$ at each timestep $t$, which makes the action space choice-based.
Nevertheless, the size of $\mathcal{A}_t$ is much larger than typical RL environments (on average 800 while in the kitchen, and increases to around 1500 as the agent gets more competent and creates new objects), and these actions are not guaranteed to have an effect to the state, which pose great challenges for RL agents to discover new experiences.

\begin{figure}[t!]
\centering
\includegraphics[width=220pt]{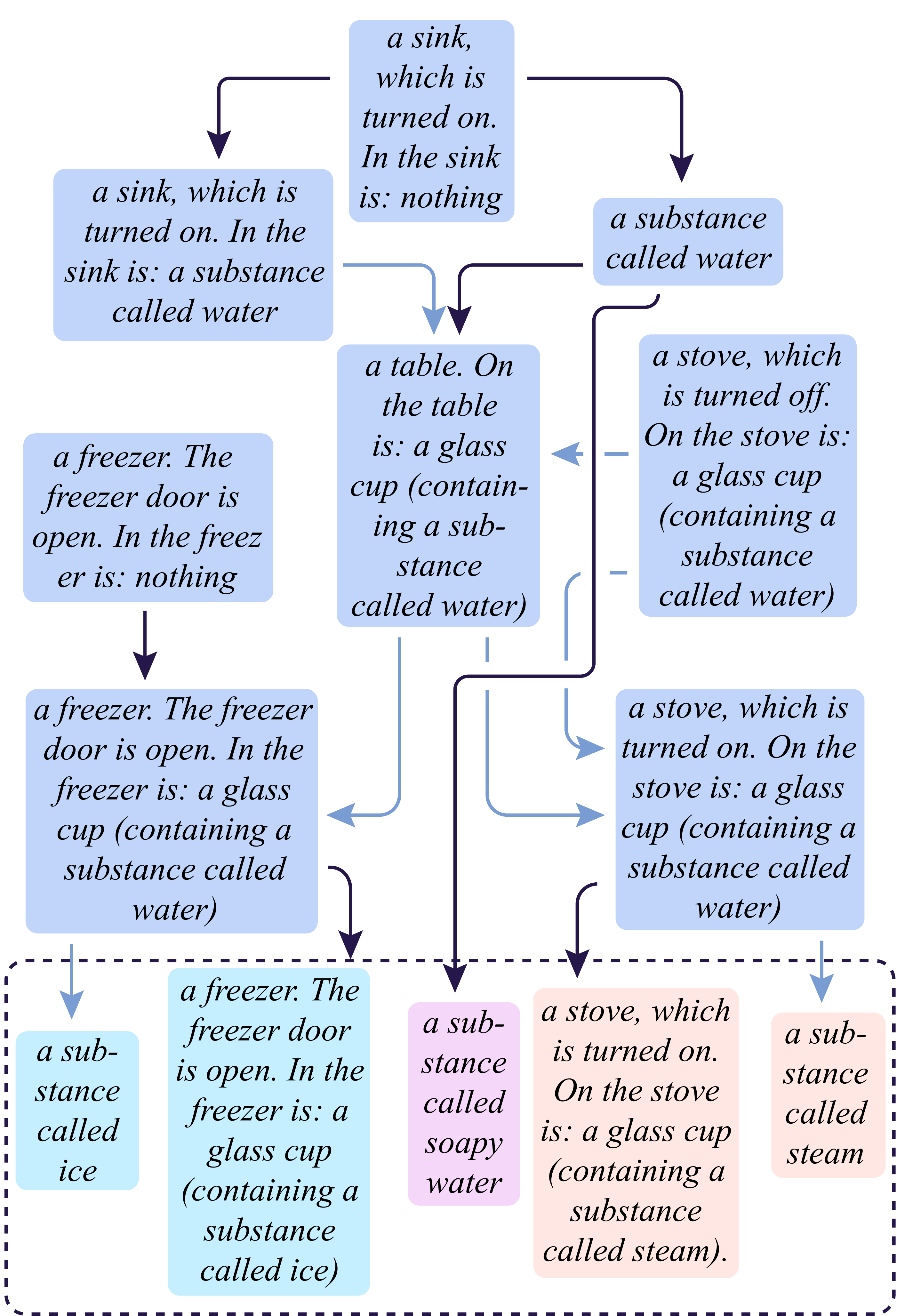}
\caption{Goal hierarchy for goals in $\mathcal{G}_{\mathcal{SP}}$. Goals are object descriptions in the environment (see main text). Light-colored arrows indicate the first goal is helpful for achieving the second one, dark-colored arrows indicate the first goal is necessary to achieve the second one. \textit{Hard goals} are the last goals in the hierarchy, inside the dashed area. For instance, if the goal does not specify steam must be made on the stove, it can be obtained in some other way.
The same goes for creating water; if the goal does not specify in which way this must be done, any container will work: either \expblue{close drain} once the \expblue{sink} is open, or \expblue{pour}ing the contents of the \expblue{table} (the \expblue{glass cup}) into the \expblue{sink}.}
\label{fig:tech_tree}
\end{figure}

We use elements in room descriptions to represent \textbf{goals}.
For instance, examples of valid goals for the trajectory shown in Figure~\ref{fig:scienceworld-trajectory-example} could be \expblue{the agent} (which is always true) or \expblue{a substance called sodium chloride}. 
This simplified goal representation facilitates relabelling and building the goal-conditioned reward function. 
As outlined in Section~\ref{sec:intro}, language-based autotelic agents require a reward function to be able to score trajectories against the original goal. 
Our uni-modal reward function operates in language space, which enables sub-string matching: a goal $g$ is valid if it can be found verbatim in the \texttt{look} feedback provided by the environment. 
While conceptually straightforward, this goal representation is rather expressive. For instance, if the goal targets an immovable object that can only be found in a certain room, this amounts to a navigation goal; if the goal targets an object that is found in a closed receptacle, accomplishing the goal requires opening the receptacle; if the goal targets an object that does not exist in the environment, then the goal amounts to making this object, which can imply a long action sequence.

\subsection{A song of ice and fire}
ScienceWorld tasks are hard exploration problems, as outlined above when considering the number of valid actions per step. In this work, we restrict ourselves to the kitchen to maintain manageable exploration, and we focus on a subset of ScienceWorld tasks: freezing and boiling water. We study the agent's progress through a self-organized curriculum of nested goals of varying difficulty, guiding the agent towards the most difficult ones. The entire tech tree and its dependency structure is presented in Figure \ref{fig:tech_tree}.

In fact, due to the non-linear design of ScienceWorld, some of the goals can be achieved without following the goal dependency structure: nevertheless this structure is useful to guide the agent's exploration. 
In this work, we will in particular look at how achieving the first, easier goals can allow the agent to master the harder ones, e.g. how the agent can create its own curriculum for learning.
Importantly, this paper is not about solving the ScienceWorld benchmark itself, but about understanding how multi-goal agents can explore their environment in the presence of large linguistic action spaces and nested goals of very different difficulty.

\section{Autotelic agents in ScienceWorld}
\label{sec:method}

\subsection{Base autotelic agent}
\label{sec:base}

Autotelic agents can be implemented with either on-policy or off-policy methods. We adopt the strongest-performing system on the original ScienceWorld benchmark, the deep reinforcement relevance network (DRRN) \cite{he2015deep}, as our internal goal-conditioned policy $\pi(.|g)$.

\label{sec:policy}
The DRRN is an off-policy deep RL agent equipped with a replay buffer. At inference time, the three observation channels (i.e., \code{obs}, \code{look} and \code{inv}) as well as the goal $g$ are tokenized using a pre-trained sentencepiece \footnote{\href{https://github.com/google/sentencepiece}{https://github.com/google/sentencepiece}} tokenizer, 
they are encoded by a GRU~\cite{cho2014gru} and the output representations are concatenated as a vector.
In parallel, all valid actions in $\mathcal{A}_t$ are tokenized and encoded with another GRU, to produce a list of valid action encodings. 
The goal-state encoding is concatenated to all valid action encodings and passed through a 1-layer MLP with ReLU activation to produce $Q(s_t, a)$ for all $a \in \mathcal{A}_t$. 
The action is either sampled from the resulting distribution (at training time) or the argmax is taken (at evaluation time). 


\textbf{Goal sampling}
At the beginning of every episode, a linguistic goal $g$ is sampled from $\mathcal{G}_{a}$, i.e. the set of goals experienced by the agent so far. In the basic case, we assume goal sampling is done uniformly.
The agent conditions its policy on that goal towards achieving it. 
An episode terminates either when the goal is achieved or after $T$ timesteps. Then, the social peer (\SP) relabels the trajectory with goals they find relevant, i.e. contained in $\mathcal{G}_\mathcal{SP}$, that were effectively achieved during that particular trajectory. The relabelling process is a discrete, linguistic version of hindsight experience replay (HER) \cite{andrychowicz2017hindsight}. The resulting relabelled trajectories, with their associated internal reward, are then pushed into the agent's replay buffer, and any goals discovered in this way are added to $\mathcal{G}_{a}$.

\textbf{Goal-modular replay buffer} 
We develop a trajectory-based, modular replay buffer.
Specifically, we store each trajectory, paired with accomplished goals (labeled by $\mathcal{SP}$) in an individual slot. 
During experience replay (in standard deep Q-learning), we use a multi-step strategy to control the transition sampling.
First, we sample a goal from the replay buffer using a certain distribution $w(g)$ (uniform unless specified otherwise). 
Given the goal, we sample a trajectory which has 0.5 probability being a positive example to the goal. 
If the sampled trajectory is a positive example, we sample a transition from it, with a probability of 0.5 that the transition has a reward.
In preliminary experiments, we observe that the above procedure (especially controlling the amount of reward the agent sees) improves sample efficiency.

\textbf{Learning}
To learn the goal-conditioned policy $\pi(.|g)$, we minimize the temporal-difference (TD) loss over transitions in the replay buffer. Given a transition $\tau_t = (s_t, a_t, s_{t+1}, r_t, \mathcal{A}_t, \mathcal{A}_{t+1})$, the TD loss is given by:
\begin{equation}
\small
TD(\tau_t) = l\big(Q(s_t, a_t), (r_t + \gamma \; \text{max}_{a' \in \mathcal{A}_{t+1}} Q(s_{t + 1}, a'))\big),
\end{equation}
where $\gamma$ is the discount factor and $r_t$ is the internal reward given by $R(s_t,a_t,g)$ when $\tau_t$ was first collected. $Q(s_t, a_t)$ is the Q-value for taking action $a_t$ in state $s_t$ and is predicted by the DRRN. The function $l$ is the smooth-L1 loss:
\begin{equation}
\small
  l(x, y) =
    \begin{cases}
      |x - y| - 0.5 & \text{if} \; |x - y| > 1;\\
      0.5 \, (x - y)^2 & \text{otherwise.}\\
    \end{cases}       
\end{equation}
To ensure exploration at training time, we add an entropy penalty term which is computed over the Q function with respect to a given $s_t$. The entropy term $H$ is also normalized by $\text{log}(|\mathcal{A}_t|)$ to account for varying numbers of valid actions across timesteps.
Therefore, the final loss is:
\begin{equation}
\small
L(\tau_t) = TD(\tau_t) + H(s_t, (a)_{a \in \mathcal{A}_t}).
\label{eq:entropy}
\end{equation}
Note there is no separate target network as no particular instability or over-optimism was found in our preliminary experiments. 



\subsection{Discovery from stepping-stone exploration}

In this section we present shortly the 2 configurations we use to study the impact of discovery from serendipity in this work: \explore and \chain. Both these mechanisms are explicitly designed to allow the agent to overcome hard-exploration problems and to master nested sets of goals, where achieving the first one is a stepping stone towards mastering the second one. One of the aims of this work is to study this effect on our ScienceWorld goals.

\textbf{\Explore}
This mechanism is very similar to the policy-based version of go-explore \cite{ecoffet2021first}. That is, after sampling a goal $g$ and the policy rollout is terminated (either by completing the goal or completing $T$ timesteps), an additional \texttt{num\_steps\_exploration} actions, set to 5 in what follows, are sampled uniformly from the set of valid actions at each timestep.

\textbf{\Chain}
This mechanism works in a similar way as \texttt{go-explore} but is more deliberate: after goal $g$ is achieved or $T$ timesteps have been achieved, with probability $p$ (0.5 in what follows) another goal $g'$ is sampled and used to condition the policy. Both \explore and \chain can be combined in a single agent.


\begin{table*}[h!]
\small
\begin{tabular}{c|l||c|c|P{2.8cm}|P{3.2cm}}
\specialrule{.15em}{.05em}{.05em} 
& Configuration Name                        & \explore                & \chain           & eval score (\textit{all}) & eval score (\textit{hard goals})    \\
\toprule\toprule
\multirow{4}{*}{\sclub} & \textbf{base}                                   & $\times$                  & $\times$        &   $71.89 \pm 16.51$     &     $50.52 \pm 47.52$      \\
& \textbf{go-explore}                             & $\checkmark$                & $\times$        &   $\boldsymbol{80.17} \pm 12.37$     &     $\boldsymbol{59.12} \pm 44.47$      \\
& \textbf{chain}                                  & $\times$                  & $\checkmark$      &   $63.60 \pm 19.48$     &     $38.24 \pm 45.01$      \\
& \textbf{go-explore-chain}                       & $\checkmark$                & $\checkmark$      &   $77.77 \pm 8.81$      &     $55.48 \pm 43.63$      \\
\midrule
\multirow{2}{*}{\sdiamond}& \textbf{no-feedback}                                  & $\times$                  & $\times$        &   $4.18 \pm 3.28$       &     $0.00 \pm 0.00$        \\
& \textbf{unconstrained} (stopped at 400k timesteps)  & $\times$                  & $\times$        &   $0.00 \pm 0.00$      &     $0.00 \pm 0.00$        \\
\midrule
\sspade & \textbf{uniform-transition}                     & $\times$                  & $\times$        &   $50.29 \pm 8.80$      &     $17.20 \pm 36.44$      \\
\midrule
\multirow{2}{*}{\sheart}& \textbf{metacognitive}                          & $\checkmark$                & $\checkmark$      &   $\boldsymbol{87.31} \pm 4.97$      &     $\boldsymbol{76.36} \pm 36.52$      \\
& \textbf{extrinsic-impossible}                   & $\times$                  & $\times$        &   $67.71 \pm 16.18$     &     $42.00 \pm 45.17$      \\
\end{tabular}
\caption{Main results of our agents on ScienceWorld. We report in the leftmost column the configuration name, in the next two if it uses \explore or \chain. The configurations are clustered by which question they answer. We then report aggregate eval scores on all our goals and aggregate eval scores on our hard goals (defined in Figure \ref{fig:tech_tree}). All eval scores are computed over the last 10 evals for stability, and are averaged across 10 random seeds. See main text for description of the configurations and commentary.}
\label{tab:main}
\end{table*}

\section{Experimental results}
\label{sec:experiment}

ScienceWorld is a challenging environment. Baseline agents are barely able to master the simplest tasks of the benchmark \cite{wang2022scienceworld}. The large action space and amount of irrelevant state changes the agent can elicit are obstacles to goal discovery, and text environments present challenges for optimization. To answer our questions, we define a set of configurations for agents that we study in what follows:
\begin{itemize}[wide, labelwidth=!, labelindent=0pt, topsep=0pt, noitemsep]
\item \textbf{base}: Our baseline agent, as described in section \ref{sec:base};
\item \textbf{go-explore}: The \textbf{base} agent equipped with \explore;
\item \textbf{chain}: The \textbf{base} agent equipped with \chain;
\item \textbf{go-explore-chain}: The base agent equipped with \explore and \chain;
\item \textbf{no-feedback}: A non-autotelic goal-conditioned agent that gets its goals uniformly from $\mathcal{G}_{\mathcal{SP}}$, and without relabeling from \SP: it only learns when stumbling upon a targeted goal by normal exploration. It is not using the goal-modular replay buffer (i.e., transition-based instead of trajectory-based).
\item \textbf{unconstrained}: A \textbf{base} agent where the \SP relabels all possible goals ($\mathcal{G}_{\mathcal{SP}} = \mathcal{G}$);
\item \textbf{uniform-transition}: A base agent, with weights $w_g$ for replay goal-sampling proportional to the total number of transitions for this goal in the replay buffer;
\item \textbf{metacognitive} An autotelic agent using both \explore and \chain, which samples its goals for an episode according to recorded intermediate competence of these goals;
\item \textbf{extrinsic-impossible}: A non-autotelic agent that gets its goals uniformly from $\mathcal{G}_{\mathcal{SP}}$, which contains an additional 100 impossible \textit{nonsense} goals.
\end{itemize}

We train our agent on 800k steps with an episode length of $T = 30$. We use as evaluation metrics the aggregate scores on all our goals as well as evaluation on \textit{hard goals}, that we define as goals where the agent needs to get some water first (see Figure \ref{fig:tech_tree}). Table \ref{tab:main} presents the results. We notice the variance is very high in all but the \textbf{metacognitive} configuration: this is due to the compounding effects of goal discovery. An agent that stumbles on the easiest goals by chance early in training will be heavily advantaged in its goal-discover compared to an agent that only sees the goal later.

\subsection{What is the role of \SP relabels in autotelic learning? (\sclub~vs \sdiamond)}
\label{sec:exp:relabel}


The dynamics of the relabelling process is of paramount importance for any learning to take place at all. In Table \ref{tab:main}, third section, we present the evaluation scores for a set of experiments with respectively an absent or a talkative \SP: in the \textbf{no-feedback} configuration the trajectories are simply input with the original goal that was targeted and the associated sparse reward; whereas for the \textbf{unconstrained} configuration, any goal that the agent accomplishes is given by the social peer as a relabel of the current trajectory ($\mathcal{G}_{\mathcal{SP}} == \mathcal{G}$). As in other configurations, goals are given if they are accomplished at any point in time. Since the goal space includes any possible descriptive changes on currently observed objects and that most actions result in such changes, the number of relabelled goals per episode in the \textbf{unconstrained} experiments is extraordinary (this agent discovers on the order of 50k goals). Both \textbf{no-feedback} and \textbf{unconstrained} configurations result in almost-null evaluation scores.

In the \textbf{no-feedback} configuration, sparseness of reward is to blame. If we let a random agent explore the room for 800k timesteps, it will encounter goals in $\mathcal{G}_{\mathcal{SP}}$ only a handful of times, and none of the hard ones (see Table \ref{tab:goal_visits}.) For the \textbf{unconstrained} configuration, there are such an important number of discovered goals (e.g., containers inside other containers inside other containers, leading to a combinatorial explosion of possible goals). This means that trajectories leading to goals in $G_{\mathcal{SP}}$ are drowned out in the set of other, non-relevant trajectories, and are only sampled a handful of times for replay; the agent's network almost never sees any reward on them, to tell nothing of the optimization process.

Lessons learned: for a multi-task agent to learn correctly in this setting, it needs to have relevant feedback from its social peer, i.e. feedback that is not too descriptive and that is relevant to the goals the social peer wants to instill in the agent.

\subsection{How to correctly prioritize hard goals in replay for an agent to be able to learn them? (\sclub~vs \sspade)}
\label{sec:exp:replay}

Another important feature of off-policy multi-goal text agents is their ability to learn a distribution of goals of varying difficulty from their replay buffer. Because the data is collected online, the replay buffer will contain many more exemplars of trajectories for the easy goals compared to the hard goals. With a vanilla replay buffer with uniform replay probabilities over transitions, the transitions corresponding to a difficult goal will hardly get replayed, drowned in the abundance of transitions from easier goals. The ratio of easy-goal transitions to hard-goal transitions only gets worse as the agent achieves mastery of the easy goals and more such transitions fill the buffer. This motivates the use of the modular goal buffer described in section \ref{sec:policy}. To empirically validate our choice, we compare the modular buffer with one mimicking the functioning of a basic replay buffer: to do so, instead of sampling goals to replay uniformly, we sample goals with a weight proportional to the number of transitions in all trajectories corresponding to this goal (the \textbf{uniform-transition} configuration). This configuration achieves lower performance and plateaus sooner compared to the \textbf{base} configuration which serves as our baseline. 


Additionally, we investigate whether other replay distribution over goals are important for learning. We investigate difficulty-based sampling (the weight of a goal is given by agent competence over the last 50 attempts of the goal). We also investigate the intermediate difficulty configuration, where the weight given to a goal in goal-sampling $w_g$ depends on empirical competence $c$ on this goal using the following formula: 
\begin{equation}
\label{eq:competence}
\small
w_{g} = f_c(c) = \alpha \; \text{exp}\Bigg(\frac{(c - 0.5)^2}{2\sigma^2}\Bigg) + \beta,
\end{equation}
$\alpha$ and $\beta$ are set to $1.0$ and $0.2$ respectively. We provide the results in the first row of the Table \ref{tab:main}. We hardly see any difference between these different goal sampling configurations.

Lessons learned: in a multitask agent with tasks of varying difficulty, where exemplars for these tasks are present at very different rates in the replay buffer, it is important to have a replay mechanism that samples often enough transitions for rare goals.

\subsection{What is the role of goal distributions when sampling goals for exploration?  (\sclub~vs \sheart)}
\label{sec:exp:goal_sampling}

We introduce the \textbf{metacognitive} agent configuration (so-called because it uses knowledge of its own competence to target goals): in this configuration the \textbf{base} agent is equipped with \explore, \chain and samples its goals based on intermediate competence (as defined in Equation \ref{eq:competence}).
The agent performs significantly better than all our other configurations, and also exhibits very low variance (as much as 3 times as low as other agents). The difference is even more apparent if we look at final performance on \textit{hard goals} compared with our \textbf{base} configuration (see Figure \ref{tab:main}, second column). For an autotelic agent, focusing on goals on which it experiences intermediate difficulty allows it to target goals on which there is good learning opportunity, as goals that are too easy are already mastered and benefit less from further exploration, whereas goals that are too hard are still unreachable for the agent. 



We finally describe experiments highlighting the interplay of hindsight relabelling with goal sampling. In the \textbf{extrinsic-impossible} configuration, the agent is given the list of 14 usual goals of interest plus an additional 100 \textit{nonsense} goals, consisting of the phrase \expblue{a substance called}~followed by various made-up words. We see that, contrary to intuition, the extrinsic-impossible configuration works rather well: the final evaluation score is similar (different with no significance) from the base configuration. This highlights a very important property of goal-exploration processes: for non-trivial exploration, a very good baseline is having random goal sampling that pushes the agent to have diverse behavior. As long as the agent discovers meaningful states in the environment and the \SP's behavior is helpful, diverse goal-conditioned behavior can be learned; the performance remains lower than in autotelic configurations nevertheless.


\begin{table}[t!]
\begin{center}
\begin{tabular}{l|r}
\specialrule{.15em}{.05em}{.05em} 
Goal                                                                   & \#occurences \\
\toprule\toprule
\rowcolor{lightblue}
\begin{tabular}{@{}l@{}} \textit{a freezer. The freezer door is open.} \\ \textit{In the freezer is: nothing.}\end{tabular} \       & 4975         \\
\begin{tabular}{@{}l@{}} \textit{a sink, which is turned on.} \\ \textit{In the sink is: nothing.}\end{tabular} \       & 4118         \\
\rowcolor{lightblue}
\textit{a substance called water}      & 473         \\
\begin{tabular}{@{}l@{}} \textit{a sink, which is turned on.} \\ \textit{In the sink is: a substance called water.}\end{tabular} \       & 68         \\
\end{tabular}
\caption{Occurrences of goals for a random agent run for 800k timesteps with environment reset every 30 timesteps, similar to the interaction setup of our \textbf{base} configuration. We record goals at each timestep as done by \SP. All omitted goals have 0 occurrences over the whole random run.}
\label{tab:goal_visits}
\end{center}
\end{table}

We see here that the dynamics of goal sampling are also important in the agent's exploration, and ultimately, learning. An agent that samples goals of intermediate competence creates its own curriculum where goals that are stepping stones are targeted first and further exploration proceeds from then on. On the other hand, an agent can be given unfeasible goals as long as they lead to diverse enough behavior.

\section{Related work}
\label{sec:related_work}

\paragraph{Autotelic agents, goal-exploration processes, novelty search} Autotelic agents were born from the study of intrinsic motivation and curiosity in humans \cite{10.3389/neuro.12.006.2007} and the application of these models in developmental robotics at first \cite{baranes2013active} and machine learning more recently \cite{forestier2022intrinsically, pmlr-v97-colas19a}. They are very close conceptually to other goal-exploration processes such as Go-explore \cite{ecoffet2021first}. The latter was developed to tackle hard-exploration challenges and stems from insights from novelty search \cite{novsearch}: exploration in environments with sparse or deceptive rewards can be driven by the search for novelty alone. The ability of autotelic agents to self-organize a curriculum \cite{ELMAN199371} of goals for training is a form of automatic curriculum learning \cite{portelas2020automatic} and has been studied by adversarial goal generation approaches \cite{florensa2018automatic, campero2020learning}.

\paragraph{Language-conditioned agents, language for goal-exploration} Building language-instructable agents has been one of the aims of AI research since its inception and is still a very active area of research today in machine learning \cite{anderson2018vision, luketina2019survey} and robotics \cite{doi:10.1146/annurev-control-101119-071628}; notable recent breakthroughs were achieved through use of large-scale pre-trained foundation models for planning \cite{saycan2022arxiv, huang2022language} and multi-modal grounding \cite{fan2022minedojo, jiang2022vima}. Language has been found to be beneficial for goal-exploration as well, by enabling abstraction \cite{mu2022improving, tam2022semantic}, combination of different abstraction levels \cite{mirchandani2021ella} and goal imagination \cite{colas2022language} supported by systematic generalization \cite{bahdanau2018systematic}. Go-explore has also been studied in the context of text environments \cite{madotto2020exploration}; albeit in very simple text environments with comparatively few valid actions compared to ScienceWorld and not in a multi-goal setting, as well as having distinct exploration and policy learning phases.

\paragraph{Interactive text environments} Text games are of particular importance to research at the intersection of RL and NLP, and thus for the study of language-informed agents. \cite{cote2019textworld} introduced TextWorld, the first such text environment, followed by IF environments \cite{hausknecht2020interactive}. These tasks are notoriously difficult for RL agents due to the large action space and abstract state space. Methods for exploration have been proposed in these contexts such as reducing the action space with LM action generation \cite{yao2020keep} or using novelty bonuses to counter deceptive rewards \cite{ammanabrolu2020avoid}. These works however did not investigate multigoal contexts (IF games being quite linear in nature) and the necessity to balance tasks of varying difficulty. ScienceWorld \cite{wang2022scienceworld} was explicitly introduced to investigate language model's abilities to act as agents in an interactive environment, but it also features more complexity and openness than other procedural text games and is thus a perfect testbed for language-based autotelic agents.

\section{Conclusion and further work}
\label{sec:conclusion}



In this work, we have presented a breakdown of the architecture of an autotelic agent, studied on a hard to explore part of the ScienceWorld text environment. Autotelic RL is a framework for implementing autonomous, open-ended, multi-task agents, and we have focused on the necessary internal components for these agents to perform efficient exploration and task learning. In particular, we have highlighted the need for a replay buffer that over-samples rare tasks and a social peer that provides appropriate interaction. This interaction comes in the form of relevant feedback of the agent's behavior but does not necessarily imply, in the case of the social peer directly giving goals to the agent, that the goals can be feasible: only that they lead to interesting interactions with the environment. The agent can shoot for the moon, all that matters is that it goes on to do something interesting and gets relevant feedback. Additionally, letting the agent sample and chain goals of intermediate competence for itself leads increased mastery of the hardest goals in ScienceWorld.


Overall, we are excited by the challenges and opportunities posed by textual autotelic agents.
We identify some important directions for future work. 
First, we only consider one environment variation; distributions of environments could be considered, and generalization could be studied: this can be challenging for current text agents.
Second, more advanced forms of automatic curriculum setting could be implemented, such as ones using learning progress to sample goals~\cite{pmlr-v97-colas19a}.
Third, goal sampling in this work has been limited to be taken from the list of achieved goals; a truly open-ended autotelic agent should be able to create its own novel goals based on previous achievements.
Last but not least, it is worth exploring to integrate a pre-trained large language model~\cite{brown2020language} into various parts of the pipeline, such that the \SP can alleviate the constraints of string-matching, and being able to imagine relevant but unseen goals, by leveraging commonsense knowledge from the language model.


\section*{Acknowledgements}

Experiments presented in this paper were carried out using the HPC resources of IDRIS under the
allocation 2022-[A0131011996] made by GENCI.

\nocite{langley00}

\bibliography{main}

\begin{thebibliography}{41}
\providecommand{\natexlab}[1]{#1}
\providecommand{\url}[1]{\texttt{#1}}
\expandafter\ifx\csname urlstyle\endcsname\relax
  \providecommand{\doi}[1]{doi: #1}\else
  \providecommand{\doi}{doi: \begingroup \urlstyle{rm}\Url}\fi

\bibitem[Ahn et~al.(2022)Ahn, Brohan, Brown, Chebotar, Cortes, David, Finn, Fu,
  Gopalakrishnan, Hausman, Herzog, Ho, Hsu, Ibarz, Ichter, Irpan, Jang, Ruano,
  Jeffrey, Jesmonth, Joshi, Julian, Kalashnikov, Kuang, Lee, Levine, Lu, Luu,
  Parada, Pastor, Quiambao, Rao, Rettinghouse, Reyes, Sermanet, Sievers, Tan,
  Toshev, Vanhoucke, Xia, Xiao, Xu, Xu, Yan, and Zeng]{saycan2022arxiv}
Ahn, M., Brohan, A., Brown, N., Chebotar, Y., Cortes, O., David, B., Finn, C.,
  Fu, C., Gopalakrishnan, K., Hausman, K., Herzog, A., Ho, D., Hsu, J., Ibarz,
  J., Ichter, B., Irpan, A., Jang, E., Ruano, R.~J., Jeffrey, K., Jesmonth, S.,
  Joshi, N., Julian, R., Kalashnikov, D., Kuang, Y., Lee, K.-H., Levine, S.,
  Lu, Y., Luu, L., Parada, C., Pastor, P., Quiambao, J., Rao, K., Rettinghouse,
  J., Reyes, D., Sermanet, P., Sievers, N., Tan, C., Toshev, A., Vanhoucke, V.,
  Xia, F., Xiao, T., Xu, P., Xu, S., Yan, M., and Zeng, A.
\newblock Do as i can and not as i say: Grounding language in robotic
  affordances.
\newblock In \emph{arXiv preprint arXiv:2204.01691}, 2022.

\bibitem[Ammanabrolu et~al.(2020)Ammanabrolu, Tien, Luo, and
  Riedl]{ammanabrolu2020avoid}
Ammanabrolu, P., Tien, E., Luo, Z., and Riedl, M.~O.
\newblock How to avoid being eaten by a grue: Exploration strategies for
  text-adventure agents.
\newblock \emph{arXiv preprint arXiv:2002.08795}, 2020.

\bibitem[Anderson et~al.(2018)Anderson, Wu, Teney, Bruce, Johnson,
  S{\"u}nderhauf, Reid, Gould, and Van Den~Hengel]{anderson2018vision}
Anderson, P., Wu, Q., Teney, D., Bruce, J., Johnson, M., S{\"u}nderhauf, N.,
  Reid, I., Gould, S., and Van Den~Hengel, A.
\newblock Vision-and-language navigation: Interpreting visually-grounded
  navigation instructions in real environments.
\newblock In \emph{Proceedings of the IEEE conference on computer vision and
  pattern recognition}, pp.\  3674--3683, 2018.

\bibitem[Andrychowicz et~al.(2017)Andrychowicz, Wolski, Ray, Schneider, Fong,
  Welinder, McGrew, Tobin, Pieter~Abbeel, and
  Zaremba]{andrychowicz2017hindsight}
Andrychowicz, M., Wolski, F., Ray, A., Schneider, J., Fong, R., Welinder, P.,
  McGrew, B., Tobin, J., Pieter~Abbeel, O., and Zaremba, W.
\newblock Hindsight experience replay.
\newblock \emph{Advances in neural information processing systems}, 30, 2017.

\bibitem[Bahdanau et~al.(2019)Bahdanau, Murty, Noukhovitch, Nguyen, de~Vries,
  and Courville]{bahdanau2018systematic}
Bahdanau, D., Murty, S., Noukhovitch, M., Nguyen, T.~H., de~Vries, H., and
  Courville, A.
\newblock Systematic generalization: What is required and can it be learned?
\newblock In \emph{International Conference on Learning Representations}, 2019.
\newblock URL \url{https://openreview.net/forum?id=HkezXnA9YX}.

\bibitem[Baranes \& Oudeyer(2013)Baranes and Oudeyer]{baranes2013active}
Baranes, A. and Oudeyer, P.-Y.
\newblock Active learning of inverse models with intrinsically motivated goal
  exploration in robots.
\newblock \emph{Robotics and Autonomous Systems}, 61\penalty0 (1):\penalty0
  49--73, 2013.

\bibitem[Bellemare et~al.(2016)Bellemare, Srinivasan, Ostrovski, Schaul,
  Saxton, and Munos]{bellemare2016unifying}
Bellemare, M., Srinivasan, S., Ostrovski, G., Schaul, T., Saxton, D., and
  Munos, R.
\newblock Unifying count-based exploration and intrinsic motivation.
\newblock \emph{Advances in neural information processing systems}, 29, 2016.

\bibitem[Brown et~al.(2020)Brown, Mann, Ryder, Subbiah, Kaplan, Dhariwal,
  Neelakantan, Shyam, Sastry, Askell, et~al.]{brown2020language}
Brown, T., Mann, B., Ryder, N., Subbiah, M., Kaplan, J.~D., Dhariwal, P.,
  Neelakantan, A., Shyam, P., Sastry, G., Askell, A., et~al.
\newblock Language models are few-shot learners.
\newblock \emph{Advances in neural information processing systems},
  33:\penalty0 1877--1901, 2020.

\bibitem[Campero et~al.(2021)Campero, Raileanu, Kuttler, Tenenbaum,
  Rockt{\"a}schel, and Grefenstette]{campero2020learning}
Campero, A., Raileanu, R., Kuttler, H., Tenenbaum, J.~B., Rockt{\"a}schel, T.,
  and Grefenstette, E.
\newblock Learning with {AMIG}o: Adversarially motivated intrinsic goals.
\newblock In \emph{International Conference on Learning Representations}, 2021.
\newblock URL \url{https://openreview.net/forum?id=ETBc_MIMgoX}.

\bibitem[Cho et~al.(2014)Cho, van Merri{\"e}nboer, Gulcehre, Bahdanau,
  Bougares, Schwenk, and Bengio]{cho2014gru}
Cho, K., van Merri{\"e}nboer, B., Gulcehre, C., Bahdanau, D., Bougares, F.,
  Schwenk, H., and Bengio, Y.
\newblock Learning phrase representations using {RNN} encoder{--}decoder for
  statistical machine translation.
\newblock In \emph{Proceedings of the 2014 Conference on Empirical Methods in
  Natural Language Processing ({EMNLP})}, pp.\  1724--1734, Doha, Qatar,
  October 2014. Association for Computational Linguistics.
\newblock \doi{10.3115/v1/D14-1179}.
\newblock URL \url{https://aclanthology.org/D14-1179}.

\bibitem[Clark(2006)]{Clark2006-CLALEA-3}
Clark, A.
\newblock Language, embodiment, and the cognitive niche.
\newblock \emph{Trends in Cognitive Sciences}, 10\penalty0 (8):\penalty0
  370--374, 2006.
\newblock \doi{10.1016/j.tics.2006.06.012}.

\bibitem[Colas et~al.(2019)Colas, Fournier, Chetouani, Sigaud, and
  Oudeyer]{pmlr-v97-colas19a}
Colas, C., Fournier, P., Chetouani, M., Sigaud, O., and Oudeyer, P.-Y.
\newblock {CURIOUS}: Intrinsically motivated modular multi-goal reinforcement
  learning.
\newblock In Chaudhuri, K. and Salakhutdinov, R. (eds.), \emph{Proceedings of
  the 36th International Conference on Machine Learning}, volume~97 of
  \emph{Proceedings of Machine Learning Research}, pp.\  1331--1340. PMLR,
  09--15 Jun 2019.
\newblock URL \url{https://proceedings.mlr.press/v97/colas19a.html}.

\bibitem[Colas et~al.(2020)Colas, Karch, Lair, Dussoux, Moulin-Frier, Dominey,
  and Oudeyer]{colas2020language}
Colas, C., Karch, T., Lair, N., Dussoux, J.-M., Moulin-Frier, C., Dominey, P.,
  and Oudeyer, P.-Y.
\newblock Language as a cognitive tool to imagine goals in curiosity driven
  exploration.
\newblock \emph{Advances in Neural Information Processing Systems},
  33:\penalty0 3761--3774, 2020.

\bibitem[Colas et~al.(2022{\natexlab{a}})Colas, Karch, Moulin-Frier, and
  Oudeyer]{colas2022language}
Colas, C., Karch, T., Moulin-Frier, C., and Oudeyer, P.-Y.
\newblock Language and culture internalization for human-like autotelic ai.
\newblock \emph{Nature Machine Intelligence}, 4\penalty0 (12):\penalty0
  1068--1076, 2022{\natexlab{a}}.

\bibitem[Colas et~al.(2022{\natexlab{b}})Colas, Karch, Sigaud, and
  Oudeyer]{colas2022autotelic}
Colas, C., Karch, T., Sigaud, O., and Oudeyer, P.-Y.
\newblock Autotelic agents with intrinsically motivated goal-conditioned
  reinforcement learning: a short survey.
\newblock \emph{Journal of Artificial Intelligence Research}, 74:\penalty0
  1159--1199, 2022{\natexlab{b}}.

\bibitem[C{\^o}t{\'e} et~al.(2019)C{\^o}t{\'e}, K{\'a}d{\'a}r, Yuan, Kybartas,
  Barnes, Fine, Moore, Hausknecht, El~Asri, Adada, et~al.]{cote2019textworld}
C{\^o}t{\'e}, M.-A., K{\'a}d{\'a}r, A., Yuan, X., Kybartas, B., Barnes, T.,
  Fine, E., Moore, J., Hausknecht, M., El~Asri, L., Adada, M., et~al.
\newblock Textworld: A learning environment for text-based games.
\newblock In \emph{Computer Games: 7th Workshop, CGW 2018, Held in Conjunction
  with the 27th International Conference on Artificial Intelligence, IJCAI
  2018, Stockholm, Sweden, July 13, 2018, Revised Selected Papers 7}, pp.\
  41--75. Springer, 2019.

\bibitem[Ecoffet et~al.(2021)Ecoffet, Huizinga, Lehman, Stanley, and
  Clune]{ecoffet2021first}
Ecoffet, A., Huizinga, J., Lehman, J., Stanley, K.~O., and Clune, J.
\newblock First return, then explore.
\newblock \emph{Nature}, 590\penalty0 (7847):\penalty0 580--586, 2021.

\bibitem[Elman(1993)]{ELMAN199371}
Elman, J.~L.
\newblock Learning and development in neural networks: the importance of
  starting small.
\newblock \emph{Cognition}, 48\penalty0 (1):\penalty0 71--99, 1993.
\newblock ISSN 0010-0277.
\newblock \doi{https://doi.org/10.1016/0010-0277(93)90058-4}.
\newblock URL
  \url{https://www.sciencedirect.com/science/article/pii/0010027793900584}.

\bibitem[Fan et~al.(2022)Fan, Wang, Jiang, Mandlekar, Yang, Zhu, Tang, Huang,
  Zhu, and Anandkumar]{fan2022minedojo}
Fan, L., Wang, G., Jiang, Y., Mandlekar, A., Yang, Y., Zhu, H., Tang, A.,
  Huang, D.-A., Zhu, Y., and Anandkumar, A.
\newblock Minedojo: Building open-ended embodied agents with internet-scale
  knowledge.
\newblock In \emph{Thirty-sixth Conference on Neural Information Processing
  Systems Datasets and Benchmarks Track}, 2022.
\newblock URL \url{https://openreview.net/forum?id=rc8o_j8I8PX}.

\bibitem[Florensa et~al.(2018)Florensa, Held, Geng, and
  Abbeel]{florensa2018automatic}
Florensa, C., Held, D., Geng, X., and Abbeel, P.
\newblock Automatic goal generation for reinforcement learning agents.
\newblock In \emph{International conference on machine learning}, pp.\
  1515--1528. PMLR, 2018.

\bibitem[Forestier et~al.(2022)Forestier, Portelas, Mollard, and
  Oudeyer]{forestier2022intrinsically}
Forestier, S., Portelas, R., Mollard, Y., and Oudeyer, P.-Y.
\newblock Intrinsically motivated goal exploration processes with automatic
  curriculum learning.
\newblock \emph{Journal of Machine Learning Research}, 2022.

\bibitem[Hausknecht et~al.(2020)Hausknecht, Ammanabrolu, C{\^o}t{\'e}, and
  Yuan]{hausknecht2020interactive}
Hausknecht, M., Ammanabrolu, P., C{\^o}t{\'e}, M.-A., and Yuan, X.
\newblock Interactive fiction games: A colossal adventure.
\newblock In \emph{Proceedings of the AAAI Conference on Artificial
  Intelligence}, volume~34, pp.\  7903--7910, 2020.

\bibitem[He et~al.(2016)He, Chen, He, Gao, Li, Deng, and Ostendorf]{he2015deep}
He, J., Chen, J., He, X., Gao, J., Li, L., Deng, L., and Ostendorf, M.
\newblock Deep reinforcement learning with a natural language action space.
\newblock In \emph{Proceedings of the 54th Annual Meeting of the Association
  for Computational Linguistics (Volume 1: Long Papers)}, pp.\  1621--1630,
  Berlin, Germany, August 2016. Association for Computational Linguistics.
\newblock \doi{10.18653/v1/P16-1153}.
\newblock URL \url{https://aclanthology.org/P16-1153}.

\bibitem[Huang et~al.(2022)Huang, Abbeel, Pathak, and
  Mordatch]{huang2022language}
Huang, W., Abbeel, P., Pathak, D., and Mordatch, I.
\newblock Language models as zero-shot planners: Extracting actionable
  knowledge for embodied agents.
\newblock \emph{arXiv preprint arXiv:2201.07207}, 2022.

\bibitem[Jansen(2022)]{jansen2022survey}
Jansen, P.
\newblock A systematic survey of text worlds as embodied natural language
  environments.
\newblock In Cote, M.-A., Yuan, X., and Ammanabrolu, P. (eds.), \emph{Wordplay
  2022 - 3rd Wordplay}, Wordplay 2022 - 3rd Wordplay: When Language Meets Games
  Workshop, Proceedings of the Workshop, pp.\  1--15. Association for
  Computational Linguistics (ACL), 2022.
\newblock Publisher Copyright: {\textcopyright} 2022 Association for
  Computational Linguistics.; 3rd Wordplay: When Language Meets Games Workshop,
  Wordplay 2022 ; Conference date: 14-07-2022.

\bibitem[Jiang et~al.(2022)Jiang, Gupta, Zhang, Wang, Dou, Chen, Fei-Fei,
  Anandkumar, Zhu, and Fan]{jiang2022vima}
Jiang, Y., Gupta, A., Zhang, Z., Wang, G., Dou, Y., Chen, Y., Fei-Fei, L.,
  Anandkumar, A., Zhu, Y., and Fan, L.
\newblock Vima: General robot manipulation with multimodal prompts.
\newblock \emph{arXiv preprint arXiv:2210.03094}, 2022.

\bibitem[Lehman \& Stanley(2011)Lehman and Stanley]{novsearch}
Lehman, J. and Stanley, K.
\newblock \emph{Novelty Search and the Problem with Objectives}, pp.\  37--56.
\newblock 11 2011.
\newblock ISBN 978-1-4614-1769-9.
\newblock \doi{10.1007/978-1-4614-1770-5_3}.

\bibitem[Luketina et~al.(2019)Luketina, Nardelli, Farquhar, Foerster, Andreas,
  Grefenstette, Whiteson, and Rockt{\"a}schel]{luketina2019survey}
Luketina, J., Nardelli, N., Farquhar, G., Foerster, J., Andreas, J.,
  Grefenstette, E., Whiteson, S., and Rockt{\"a}schel, T.
\newblock A survey of reinforcement learning informed by natural language.
\newblock \emph{IJCAI}, 2019.

\bibitem[Madotto et~al.(2021)Madotto, Namazifar, Huizinga, Molino, Ecoffet,
  Zheng, Yu, Papangelis, Khatri, and Tur]{madotto2020exploration}
Madotto, A., Namazifar, M., Huizinga, J., Molino, P., Ecoffet, A., Zheng, H.,
  Yu, D., Papangelis, A., Khatri, C., and Tur, G.
\newblock Exploration based language learning for text-based games.
\newblock In \emph{Proceedings of the Twenty-Ninth International Joint
  Conference on Artificial Intelligence}, IJCAI'20, 2021.
\newblock ISBN 9780999241165.

\bibitem[Mirchandani et~al.(2021)Mirchandani, Karamcheti, and
  Sadigh]{mirchandani2021ella}
Mirchandani, S., Karamcheti, S., and Sadigh, D.
\newblock Ella: Exploration through learned language abstraction.
\newblock \emph{Advances in Neural Information Processing Systems},
  34:\penalty0 29529--29540, 2021.

\bibitem[Mu et~al.(2022)Mu, Zhong, Raileanu, Jiang, Goodman, Rockt{\"a}schel,
  and Grefenstette]{mu2022improving}
Mu, J., Zhong, V., Raileanu, R., Jiang, M., Goodman, N., Rockt{\"a}schel, T.,
  and Grefenstette, E.
\newblock Improving intrinsic exploration with language abstractions.
\newblock In Oh, A.~H., Agarwal, A., Belgrave, D., and Cho, K. (eds.),
  \emph{Advances in Neural Information Processing Systems}, 2022.
\newblock URL \url{https://openreview.net/forum?id=ALIYCycCsTy}.

\bibitem[Oudeyer \& Kaplan(2007)Oudeyer and Kaplan]{10.3389/neuro.12.006.2007}
Oudeyer, P.-Y. and Kaplan, F.
\newblock What is intrinsic motivation? a typology of computational approaches.
\newblock \emph{Frontiers in Neurorobotics}, 1, 2007.
\newblock ISSN 1662-5218.
\newblock \doi{10.3389/neuro.12.006.2007}.
\newblock URL
  \url{https://www.frontiersin.org/articles/10.3389/neuro.12.006.2007}.

\bibitem[Portelas et~al.(2021)Portelas, Colas, Weng, Hofmann, and
  Oudeyer]{portelas2020automatic}
Portelas, R., Colas, C., Weng, L., Hofmann, K., and Oudeyer, P.-Y.
\newblock Automatic curriculum learning for deep rl: A short survey.
\newblock In \emph{Proceedings of the Twenty-Ninth International Joint
  Conference on Artificial Intelligence}, IJCAI'20, 2021.
\newblock ISBN 9780999241165.

\bibitem[Péré et~al.(2018)Péré, Forestier, Sigaud, and
  Oudeyer]{pere2018unsupervised}
Péré, A., Forestier, S., Sigaud, O., and Oudeyer, P.-Y.
\newblock Unsupervised learning of goal spaces for intrinsically motivated goal
  exploration.
\newblock In \emph{International Conference on Learning Representations}, 2018.
\newblock URL \url{https://openreview.net/forum?id=S1DWPP1A-}.

\bibitem[Shridhar et~al.(2020)Shridhar, Yuan, C{\^o}t{\'e}, Bisk, Trischler,
  and Hausknecht]{shridhar2020alfworld}
Shridhar, M., Yuan, X., C{\^o}t{\'e}, M.-A., Bisk, Y., Trischler, A., and
  Hausknecht, M.
\newblock Alfworld: Aligning text and embodied environments for interactive
  learning.
\newblock \emph{arXiv preprint arXiv:2010.03768}, 2020.

\bibitem[Sigaud et~al.(2021)Sigaud, Caselles-Dupr{\'e}, Colas, Akakzia,
  Oudeyer, and Chetouani]{sigaud2021towards}
Sigaud, O., Caselles-Dupr{\'e}, H., Colas, C., Akakzia, A., Oudeyer, P.-Y., and
  Chetouani, M.
\newblock Towards teachable autonomous agents.
\newblock \emph{arXiv preprint arXiv:2105.11977}, 2021.

\bibitem[Sutton \& Barto(2018)Sutton and Barto]{SuttonRL}
Sutton, R.~S. and Barto, A.~G.
\newblock \emph{Reinforcement Learning: An Introduction}.
\newblock The MIT Press, 2018.

\bibitem[Tam et~al.(2022)Tam, Rabinowitz, Lampinen, Roy, Chan, Strouse, Wang,
  Banino, and Hill]{tam2022semantic}
Tam, A., Rabinowitz, N.~C., Lampinen, A.~K., Roy, N.~A., Chan, S.~C., Strouse,
  D., Wang, J.~X., Banino, A., and Hill, F.
\newblock Semantic exploration from language abstractions and pretrained
  representations.
\newblock In Oh, A.~H., Agarwal, A., Belgrave, D., and Cho, K. (eds.),
  \emph{Advances in Neural Information Processing Systems}, 2022.
\newblock URL \url{https://openreview.net/forum?id=-NOQJw5z_KY}.

\bibitem[Tellex et~al.(2020)Tellex, Gopalan, Kress-Gazit, and
  Matuszek]{doi:10.1146/annurev-control-101119-071628}
Tellex, S., Gopalan, N., Kress-Gazit, H., and Matuszek, C.
\newblock Robots that use language.
\newblock \emph{Annual Review of Control, Robotics, and Autonomous Systems},
  3\penalty0 (1):\penalty0 25--55, 2020.
\newblock \doi{10.1146/annurev-control-101119-071628}.
\newblock URL \url{https://doi.org/10.1146/annurev-control-101119-071628}.

\bibitem[Wang et~al.(2022)Wang, Jansen, C{\^o}t{\'e}, and
  Ammanabrolu]{wang2022scienceworld}
Wang, R., Jansen, P., C{\^o}t{\'e}, M.-A., and Ammanabrolu, P.
\newblock Scienceworld: Is your agent smarter than a 5th grader?
\newblock In \emph{Proceedings of the 2022 Conference on Empirical Methods in
  Natural Language Processing (EMNLP)}. Association for Computational
  Linguistics, December 2022.

\bibitem[Yao et~al.(2020)Yao, Rao, Hausknecht, and Narasimhan]{yao2020keep}
Yao, S., Rao, R., Hausknecht, M., and Narasimhan, K.
\newblock Keep {CALM} and explore: Language models for action generation in
  text-based games.
\newblock In \emph{Proceedings of the 2020 Conference on Empirical Methods in
  Natural Language Processing (EMNLP)}, pp.\  8736--8754, Online, November
  2020. Association for Computational Linguistics.
\newblock \doi{10.18653/v1/2020.emnlp-main.704}.
\newblock URL \url{https://aclanthology.org/2020.emnlp-main.704}.

\end{thebibliography}
\bibliographystyle{icml2023}

\label{asec:examples}

\end{document}